# A Noise Resilient Approach for Robust Hurst Exponent Estimation

Malith Premarathna

Department of Statistics, University of Nebraska-Lincoln,

Fabrizio Ruggeri

CNR IMATI, Milano

and

Dixon Vimalajeewa

Department of Statistics, University of Nebraska-Lincoln

October 3, 2025

**Abstract**

Understanding signal behavior across scales is vital in areas such as natural phenomena analysis and financial modeling. A key property is self-similarity, quantified by the Hurst exponent ($H$), which reveals long-term dependencies. Wavelet-based methods are effective for estimating $H$ due to their multi-scale analysis capability, but additive noise in real-world measurements often degrades accuracy.

We propose *Noise-Controlled ALPHEE (NC-ALPHEE)*, an enhancement of the *Average Level-Pairwise Hurst Exponent Estimator (ALPHEE)*, incorporating noise mitigation and generating multiple level-pairwise estimates from signal energy pairs. A neural network (NN) combines these estimates, replacing traditional averaging. This adaptive learning maintains ALPHEE's behavior in noise-free cases while improving performance in noisy conditions.

Extensive simulations show that in noise-free data, NC-ALPHEE matches ALPHEE's accuracy using both averaging and NN-based methods. Under noise, however, traditional averaging deteriorates and requires impractical level restrictions, while NC-ALPHEE consistently outperforms existing techniques without such constraints.

NC-ALPHEE offers a robust, adaptive approach for $H$ estimation, significantly enhancing the reliability of wavelet-based methods in noisy environments.

# 1   Introduction

Recent advances in technologies such as the Internet of Things (IoT) and nanotechnology have enabled the collection of high-frequency signals from previously inaccessible environments, including in-body monitoring systemsVimalajeewa & Balasubramaniam (2021). A notable property of such signals is *self-similarity*, where complex patterns recur across multiple resolutions or scales. In statistical terms, self-similarity implies that certain characteristics of a signal—such as mean and variance—remain consistent across temporal or spatial scales. This phenomenon reflects long-range correlations in the data and is instrumental in understanding the dynamics of complex systems across diverse scientific fields. Applications include modeling ozone concentration in climatology Katul et al. (2006), investigating neural processing in neurology Campbell & Weber (2022), diagnosing diseases in medicine Vimalajeewa et al. (2025), and analyzing stock market fluctuations in finance Gómez-Águila et al. (2022).

The Hurst exponent ($H \in [0,1]$) is a key parameter used to quantify self-similarity Abry & Veitch (1998). It captures the degree of long-range dependence in a signal and provides insights into its temporal structure. Several techniques have been developed for estimating $H$, including rescaled range analysis, detrended fluctuation analysis, and power spectral density methods Raubitzek et al. (2023). Among these, wavelet-based approaches are particularly effective in handling the complexity of real-world signals. These methods estimate $H$ by analyzing the linear decay of signal energy across wavelet decomposition levels, as represented in the wavelet spectrum Vidakovic (1999). However, their performance is often compromised in the presence of noise and outliers.

In practical scenarios, signals are frequently corrupted by noise, which significantly undermines the reliability of self-similarity assessments. For instance, in cases where a self-similar signal is contaminated with additive Gaussian noise—a common occurrence in biomedical signal processing, geophysics, and finance—the observed data becomes a combination of structured long-memory components and unstructured random noise. This contamination introduces bias, distorts patterns, and may even produce artificial discontinuities that complicate self-similarity analysis Chigansky & Kleptsyna (2023), Chandrasekaran et al. (2019), Xu et al. (2025), Baglaeva et al. (2023). Addressing these challenges requires robust methods capable of mitigating the effects of noise when estimating $H$.

Several wavelet-based approaches have attempted to improve noise resilience. For example, the method proposed in Vimalajeewa, Bruce & Vidakovic (2023) introduced a distance-variance strategy to enhance estimation reliability but relied on the assumption of linear spectral decay, which does not always hold in noisy conditions. Another study Vimalajeewa, McDonald, Tung & Vidakovic (2023) computed multiple $H$ values from linear regions of the wavelet spectrum, but this approach underutilizes the available signal information. To overcome these limitations, the ALPHEE (Average Level Pairwise Hurst Exponent Estimator) method was proposed by Vimalajeewa et al. (2025). Unlike earlier techniques, ALPHEE uses energy pairs across decomposition levels to estimate $H$. While promising, ALPHEE remains vulnerable to noise contamination.

To address these challenges, we propose a novel, noise-aware approach: *Noise-Controlled ALPHEE (NC-ALPHEE)*. This method is a refined version of ALPHEE designed to improve robustness against noise in estimating the Hurst exponent in the wavelet domain. Similar to the ALPHEE method, NC-ALPHEE leverages localized distributional properties of wavelet coefficients across selected decomposition levels, eliminating the restrictive assumption of linear energy decay. The key difference from the ALPHEE is that NC-ALPHEE exploits the fact that the rescaled level-wise wavelet energy of a Gaussian self-similar process with additive noise follows a chi-squared distribution. We derive an estimator for $H$ and its variance and use the variances to compute weights for aggregating (i.e., computing weighted average of) level-pairwise estimates. At higher noise levels, however, this weighted averaging approach becomes suboptimal. To overcome this, NC-ALPHEE incorporates a neural network-based aggregation strategy that learns to combine level-pairwise estimates optimally, resulting in a more accurate and noise-resilient final estimate.

The effectiveness of NC-ALPHEE is demonstrated through extensive simulation studies. Our results show that NC-ALPHEE, augmented with neural network integration, significantly outperforms traditional wavelet spectrum-based methods and the original ALPHEE approach in assessing self-similarity under noisy conditions.

The remainder of this paper is organized as follows: Section 2 reviews the fundamentals of wavelet transforms and the background for measuring self-similarity in signals. Section 3 details the proposed NC-ALPHEE method. Section 4 describes the simulation framework used to evaluate NC-ALPHEE's performance. Section 5 presents the results, which are discussed in Section 6. Finally, Section 7 provides concluding remarks.

## 2 Wavelet Transform-based Self-Similarity Analysis

Wavelet transforms are widely used in signal and image processing. This section provides an overview of wavelet transforms, self-similarity, and the wavelet spectrum to establish a foundational understanding of wavelet-based self-similarity assessment.

This manuscript follows a standard convention in statistical physics and time series analysis: we use $H$ for the actual or theoretical Hurst exponent (for example, in simulations), and $/hatH$ for its estimated value. It is essential to make this distinction when comparing methods or evaluating the accuracy of estimations.

## 2.1 Wavelet Transforms

In signal processing, wavelet transformations (WTs) are widely used to analyze complex signals, particularly those with high-frequency components such as time series data. WT allows a signal to be broken down into components that are localized in both time and frequency, making it possible to analyze signal features at different resolutions simultaneously.

A commonly used form of wavelet transformation is the Discrete Wavelet Transform (DWT). It is especially useful in domains where data are available in discrete form. DWTs are linear and orthogonal transformations and can be expressed through simple matrix multiplication. Given a signal $Y$ of size $N \times 1$, its DWT, denoted by $d_{N \times 1}$, can be written as:

$$d = WY, \qquad (1)$$

where $W$ is an $N \times N$ orthogonal matrix determined by the chosen wavelet basis such as *Haar*, *Daubechies*, or *Symmlet* wavelets. Although $N$ can be arbitrary, it is often chosen to be a power of two, i.e., $N = 2^J$, where $J \in \mathbb{Z}^+$, to facilitate efficient computation.

To address computational challenges in evaluating (1) for large $N$, Mallat (1989) introduced a fast algorithm based on filtering. This algorithm performs a series of convolutions using a *low-pass filter* $h$ and its mirror counterpart, a *high-pass filter* $g$, both specific to the selected wavelet basis. After each convolution, the result is downsampled (i.e., every second value is retained), resulting in a multiresolution representation of the original signal.

This process yields: (1) A smoothing approximation, representing the overall trend in the signal (denoted by coefficients $c$), and (2) a hierarchy of detail wavelet coefficients $d_{jk}$ that capture high-frequency details at multiple scales. These coefficients are indexed by the scale level $j$, which indicates resolution, and the position $k$, which indicates location within the scale.

The convolution and downsampling steps are repeated until a desired decomposition level $j = J_0$ is reached, where $0 \leq J_0 \leq J - 1$ and $J = \log_2(N)$. The structure of the resulting DWT vector $d$ is:

$$d = (c_{J_0}, d_{J_0}, d_{J_0+1}, \ldots, d_{J-1}), \tag{2}$$

where $c_{J_0}$ is the approximation (low-pass or smoothing) coefficient vector of size $(2^{J_0} \times 1)$ and each $d_j$ is a vector of detail coefficients of size $2^j \times 1$, representing higher-frequency components of the signal, where $j = J_0, J_0 + 1, \cdots, J - 2, J - 1$.

As we will demonstrate later, these wavelet coefficients are powerful tools for capturing the underlying structure and dynamics of complex signal patterns that are often missed by traditional statistical analysis techniques.

## 2.2 Self-similarity

A deterministic function $f(t)$, where $t$ is a $d$-dimensional input, is said to be self-similar if there exists a constant exponent $H$ (known as the Hurst exponent) such that for any dilation factor $\lambda > 0$, the following condition holds:

$$f(\lambda t) = \lambda^{-H} f(t).$$

This concept has been extended to random processes. A stochastic process $\{X(t), t \in \mathbb{R}^d\}$ is self-similar with scaling parameter $H$ if, for any $\lambda > 0$,

$$X(\lambda t) \stackrel{d}{=} \lambda^H X(t), \tag{3}$$

where $\stackrel{d}{=}$ denotes equality in all finite-dimensional distributions.

The parameter $H$ is statistically well-defined in models that assume stationary increments and Gaussianity. Among various techniques, wavelet-based methods have become a popular choice for estimating $H$, particularly in signals that exhibit self-similar behavior. These methods rely on analyzing the wavelet spectrum, which reflects how the signal's energy varies across scales or resolutions Abry & Veitch (1998).

From the self-similarity definition in equation (3), the detail wavelet coefficients $d_{jk}$ of a self-similar process satisfy:

$$d_{jk} \stackrel{d}{=} 2^{-j(H+1/2)} d_{0k}, \tag{4}$$

for a given resolution level $j$, assuming the wavelet basis is $L_2$-normalized, where $d_{0k}$ are finest level of detail wavelet coefficients.

If the process $X(t)$ is zero-mean (i.e., $\mathbb{E}[X(t)] = 0$) and has stationary increments (i.e., the distribution of $X(t+h) - X(t)$ is independent of $t$), then $\mathbb{E}[d_{0k}] = 0$ and $\mathbb{E}[d_{0k}^2] = \mathbb{E}[d_{00}^2]$. As a result, the expected value of the squared wavelet coefficients at scale $j$ is given by:

$$\mathbb{E}[d_{jk}^2] \propto 2^{-j(2H+1)}. \tag{5}$$

### 2.2.1 The Standard Method for Estimating $H$

This method involves computing the wavelet spectrum that represents the change in signal energy across different resolutions.

Equation (5) forms the theoretical foundation for estimating the Hurst exponent. Taking the base-2 logarithm transforms this into a linear relationship:

$$\log_2 \mathbb{E}[d_{jk}^2] = \beta_0 - j\beta_1,$$

where $\beta_0$ and $\beta_1 (= 2H + 1)$ are parameters to be estimated.

The set of pairs $(j, S(j)) = \left(j, \log_2 \mathbb{E}[d_{jk}^2]\right)$ defines the wavelet spectrum, where $S(j)$ represents the log-energy at scale $j$. In practice, $S(j)$ is computed by averaging the squared wavelet coefficients at each scale and then taking the logarithm:

$$S(j) = \log_2 \left( \frac{1}{n_j} \sum_{k=1}^{n_j} d_{jk}^2 \right),$$

where $n_j$ is the number of wavelet coefficients at scale $j$.

This log-energy spectrum is expected to follow a linear trend across scales for self-similar processes. The slope $\beta_1$ of this line provides a direct means to estimate the Hurst exponent via:

$$H = \frac{\beta_1 - 1}{2}.$$

However, this method faces two key challenges when applied to real-world signals. First, the energy of these signals often deviates from the expected linear relationship. Second, real-world signals are frequently contaminated by measurement and environmental noise, which distorts the calculation of signal energy. Together, these issues hinder the accurate estimation of the Hurst exponent ($H$), underscoring the need for more robust estimation techniques. To address this, we introduce a new estimator—NC-ALPHEE—designed to provide accurate H estimates even in the presence of significant noise.

## 3 Noise Controlled Average Level Pairwise Hurst Exponent Estimator (NC-ALPHEE)

Suppose the observed signal, $Y$, is a mixture of the self-similar process $X$ and additive random noise $\epsilon$ as follows:

$$Y(t) = X(t) + \epsilon(t), t = 1, 2, \cdots N, \tag{6}$$

where $X$ is a self-similar process with Hurst exponent $H$, scaling parameter $\sigma_X^2$, with zero-mean i.e., $\mathbb{E}(X) = 0$), stationary increments (i.e., the distribution of $X(t+h) - X(h)$ is independent of $t$), and finite variance (i.e., $Var(X) = \sigma_X^2 t^{2H}$). The error $\epsilon(t)$ is assumed to be Gaussian noise with zero mean and variance $\sigma_\epsilon^2$ (i.e., $\epsilon \sim \mathcal{N}(0, \sigma_\epsilon^2)$) with no memory or correlation across time and also independent from $X$.

The estimation process assumes that the wavelet coefficients in a level $j$ are approximately independent due to their assumed Gaussianity and the decorrelation property. More specifically, the authors of Flandrin (1992) and Tewfik & Kim (1992) have discussed the decorrelation property as follows. As the distance between two wavelet coefficients increases, the covariance between them decays, and the rate of decay depends on $H$ and the number of vanishing moments $M$ of the wavelet basis used as

$$\mathbb{E} d_{jk_1} d_{jk_2} \leq C |k_1 - k_2|^{2(H-M)}, \tag{7}$$

where $C$ depends only on the level $j$. For adjacent coefficients (i.e., small $|k_1 - k_2|$), this covariance may not be small, but it exponentially decays to zero as long as $M > H$ and $|k_1 - k_2| > 1$.

To ensure short memory of $d_{jk}, k \in \mathbb{Z}^+$, the convergence of $\sum_{k_1, k_2} \mathbb{E}|d_{jk_1} d_{jk_2}|$ is needed. For this convergence, it is required that $M > H + 1/2$. Since $M = 1$ for the Haar wavelet, this wavelet basis should not be used due to poor decorrelation property.

Next, the probability distribution of wavelet coefficients of the noisy signal $Y$ given in (6) is explored at different levels.

## 3.1 Distribution of Wavelet Coefficients

Assuming the wavelet coefficients at level $j$ follow a Gaussian distribution, it follows

$$d_{jk} \sim \mathcal{N}\left(0, \sigma_{j,X}^2 + \sigma_{j,\epsilon}^2\right),$$

where $\sigma_{j,X}^2 = \sigma_X^2 2^{-j(2H+1)}$ and $\sigma_{j,\epsilon}^2 = \sigma_\epsilon^2$.

According to equation (5), if $d_{0k} \sim \mathcal{N}(0, \sigma_j^2)$, then a properly scaled squared wavelet coefficient at level $j$ follows a chi-squared distribution with one degree of freedom. That is,

$$\frac{d_{jk}^2}{\sigma_X^2 2^{-j(2H+1)} + \sigma_\epsilon^2} \sim \chi_1^2. \tag{8}$$

Following the decorrelation property of wavelet coefficients described in (7), all ($n_j = 2^j$) wavelet coefficients $\{d_{jk}\}_{k=1}^n$ at level $j$ are independent Gaussian random variables with zero mean and variance $\sigma_{jk}^2$. The squared sum of these scaled coefficients has a chi-squared distribution with $n_j$ degrees of freedom.

$$\frac{\sum_{k=1}^n d_{jk}^2}{\sigma_X^2 2^{-j(2H+1)} + \sigma_\epsilon^2} \sim \chi_{n_j}^2 \tag{9}$$

The log of (9) has a log-chi-squared distribution as follows

$$Z = \log\left(\frac{\sum_{k=1}^{n_j} d_{jk}^2}{\sigma_X^2 2^{-j(2H+1)} + \sigma_\epsilon^2}\right) \sim \log \chi_{n_j}^2 \tag{10}$$

According to the properties of the well-known log-chi-squared distribution, if $W \sim \log \chi_n^2$, then mean and variance of $W$ are given by

$$\mathbb{E}(W) = \log(2) + \psi(n/2) \quad \text{and} \quad Var(W) = \psi'(n/2), \tag{11}$$

where $\psi$ and $\psi'$ are digamma and trigamma functions, respectively.

In the following two sections, equations (10) and (11) are used together to derive the expectation and variance of the Hurst exponent estimator denoted as $\hat{H}$.

## 3.2 Expectation of $H$

Following equation (10) and (11), expected value of $Z$ can be expressed as

$$\mathbb{E}(Z) = \mathbb{E}\left[\log\left(\frac{n\overline{d_j^2}}{\sigma_X^2 2^{-j(2H+1)} + \sigma_\epsilon^2}\right)\right] = \log 2 + \psi(n_j/2).$$

This can be simplified to

$$\log n_j + \mathbb{E}[\log(\overline{d_j^2})] - \log(\sigma_X^2 2^{-j(2H+1)} + \sigma_\epsilon^2) = \log 2 + \psi(n_j/2). \tag{12}$$

In order to find an estimator for $H$, we take into account the relationship shown in (12) at two different levels. To do that, we replace $n_j$ and $j$ in (12) with $n_{j_i}$ and $j_i$ as:

$$\log n_{j_i} + \mathbb{E}\left(\log(\overline{d_{j_i}^2})\right) - \log\left(\sigma_X^2 2^{-j_i(2H+1)} + \sigma_\epsilon^2\right) = \log 2 + \psi\left(\frac{n_{j_i}}{2}\right), \quad i = 1, 2, \tag{13}$$

where $n_{j_i} = 2^{j_i}$ is the number of wavelet coefficients at level $j_i$.

Consider the relationship in (13) at two different levels $j_1$ and $j_2$. First, their difference is taken to remove an unknown parameter $\sigma_X^2$. Second, in the resulting expression, using the method of moments, the first moment of the population $\mathbb{E}(\overline{d_{j_i}^2})$ is replaced by the first moment of the sample of the wavelet coefficients $\log \overline{d_{j_i}^2}$. Finally, the resulting expression for $H$ is as follows (detailed derivation is included in Appendix).

$$\hat{H} = \frac{1}{2(j_1 - j_2)}\left[\left(\frac{\psi(n_{j_1}/2) - \psi(n_{j_2}/2)}{\log(2)}\right) - \log_2\left(\frac{n_{j_1}\overline{d_{j_1}^2} - 2\sigma_\epsilon^2 e^{\psi(n_{j_1}/2)}}{n_{j_2}\overline{d_{j_2}^2} - 2\sigma_\epsilon^2 e^{\psi(n_{j_2}/2)}}\right)\right] - \frac{1}{2}, \tag{14}$$

where $0 \leq j_1 < j_2 \leq J - 1$ and $J = \log_2 N$, with $N$ being the length of the signal.

## 3.3 Variance of $\hat{H}$ ( $V(\hat{H})$)

We use the expression in (14) along with the variance of a log-chi-squared random variable given in (11) to find the variance of the estimator $\hat{H}$.

$$V(\hat{H}) = \frac{1}{4(j_1 - j_2)^2}\left[V\left(\log_2(n_{j_1}\overline{d_{j_1}^2} - 2\sigma_\epsilon^2 e^{\psi(n_{j_1}/2)})\right) + V\left(\log_2(n_{j_2}\overline{d_{j_2}^2} - 2\sigma_\epsilon^2 e^{\psi(n_{j_2}/2)})\right)\right] \quad (15)$$

This variance can be expressed in the following equation. The details about the derivation can be found in the Appendix.

$$V(\hat{H}) \approx \frac{1}{\left(2(j_1 - j_2)\log(2)\right)^2} \sum_{i=1}^{2}\left[\psi'\left(\frac{n_{j_i}}{2}\right) + \frac{8\sigma_\epsilon^4 e^{2\psi(n_{j_i}/2)}}{\sigma_{j_i}^4(n_{j_i} - 2)^2(n_{j_i} - 4)} + \frac{8\sigma_\epsilon^2 e^{\psi(n_{j_i}/2)}}{\sigma_{j_i}^2(n_{j_i} - 2)n_{j_i}}\right] \quad (16)$$

To estimate $\hat{H}$ and $V(\hat{H})$, we need the noise variance $\sigma_\epsilon^2$ and variance of detail wavelet coefficients at decomposition level $j_i$, $\sigma_{j_i}^2$. According to Vidakovic (1999), the variance of the finest level (i.e., $j = J - 1$) of detail wavelet coefficients is a good estimator of the noise variance Vimalajeewa, DasGupta, Ruggeri & Vidakovic (2023).

$$\sigma_\epsilon^2 \approx V(d_{jk}), \quad \text{where} \quad j = J - 1.$$

The $\sigma_{j_i}^2$ is estimated as the average square of the detail wavelet coefficients at the $j_i$th level.

$$\sigma_{j_i}^2 = V(d_{j_i}) = \mathbb{E}(d_{j_i}^2) - [\mathbb{E}(d_{j_i})]^2 = \mathbb{E}(d_{j_i}^2) \approx \overline{d_{j_i}^2} \quad (\because \mathbb{E}(d_{j_i}) = 0)$$

## 3.4 Derivation of ALPHEE (NC-ALPHEE when $\sigma_\epsilon = 0$)

The ALPHEE method proposed in Vimalajeewa et al. (2025) is a special case of NC-ALPHEE when the noise variance $\sigma_\epsilon^2 = 0$.

1. **Estimation of $\hat{H}$:** The substitution of $\sigma_\epsilon^2 = 0$ in equation (14) and then conversion of $\log(2)$ to $\log_2(e)$ result in

$$\hat{H} = \frac{1}{2(j_1 - j_2)} \left[ \log_2(e) \left( \psi(n_{j_1}/2) - \psi(n_{j_2}/2) \right) - \left( \log_2 \overline{d_{j_1}^2} - \log_2 \overline{d_{j_2}^2} \right) \right] - 1 \quad (17)$$

2. **Estimation of** $V(\hat{H})$**:** Substituting $\sigma_\epsilon^2 = 0$ in (16) and taking its variance, the variance of $\hat{H}$ in the absence of noise can be expressed as

$$Var(\hat{H}) = \frac{1}{[2(j_1 - j_2)\log(2)]^2} \left[ \psi'\left(\frac{n_{j_1}}{2}\right) + \psi'\left(\frac{n_{j_2}}{2}\right) \right] \quad (18)$$

The NC-ALPHEE method produces multiple candidate estimates ($\hat{H}$s). For a signal of length $2^J$, where $J = \log_2(N)$, up to $J - 1$ wavelet decompositions can be performed, yielding $\binom{J-1}{2}$ level-pairwise estimates. These estimates must be combined effectively to produce a final estimate. The next section outlines common averaging techniques and a neural network-based method to enhance estimation accuracy.

### 3.5 Aggregation of level-pairwise estimators

Let $\hat{H}_{j_1,j_2}$ represent a candidate estimator of the Hurst exponent $H$, as defined in equation (14), computed from wavelet scales (or level pair) $j_1$ and $j_2$, where $j_1 < j_2$. An *omnibus estimator* of $H$ is proposed, constructed by aggregating all pairwise estimates $\hat{H}_{j_1,j_2}$ across a specified range of levels.

Two widely adopted aggregation methods are the *weighted mean* and the more robust *weighted median*. In both approaches, the weights $w_{j_1,j_2}$ are defined to be inversely proportional to the variance of the corresponding pairwise estimates:

$$w_{j_1,j_2} \propto \left( \text{Var}(\hat{H}_{j_1,j_2}) \right)^{-1}, \quad (19)$$

and are normalized to satisfy:

$$\sum_{j_1 < j_2} w_{j_1,j_2} = 1.$$

This weighting scheme prioritizes more stable estimates by assigning greater importance to those with lower variability, while reducing the influence of highly variable (and thus less reliable) estimates.

### 3.5.1 Weighted Mean Estimator

The estimator based on the weighted mean is defined as:

$$\hat{H} = \sum_{j_1 < j_2} w_{j_1,j_2} \hat{H}_{j_1,j_2}. \tag{20}$$

### 3.5.2 Weighted Median Estimator

The weighted median $\hat{H}_{(k)}$ is defined such that:

$$\sum_{i=1}^{k-1} w_i \leq \frac{1}{2} \quad \text{and} \quad \sum_{i=k+1}^{n} w_i \leq \frac{1}{2}, \tag{21}$$

where $\hat{H}_{(i)}$ denotes the $i$-th ordered pairwise estimator in the ensemble $\{\hat{H}_{j_1,j_2} : j_1 < j_2\}$, and $w_i$ is the corresponding weight.

These aggregation methods exhibit reduced accuracy in estimating the Hurst exponent ($H$) under conditions of significant signal contamination. As an alternative, we propose the NN method.

### 3.5.3 Neural Network Estimator

We utilize all available candidate estimators computed from every possible level pair and input them into a fully connected neural network (NN). This network consists of an input layer with a number of nodes equal to the total number of candidate estimators, a tunable number of hidden layers (ranging from two to five), and a single-node output layer. The number of nodes in each hidden layer, the activation function, learning rate, batch size, and weight decay are treated as tunable parameters. This NN architecture is optimized using the Optuna framework (Akiba et al. (2019)), which performs a Bayesian search over the hyperparameter space to identify configurations that minimize the mean squared error (MSE). A five-fold cross-validation scheme is employed to ensure robust generalization, and early stopping is incorporated within each fold to prevent overfitting and reduce computational cost. The final model balances flexibility and efficiency, enabling accurate approximation of H while avoiding excessive complexity.

## 4  Performance Evaluation

This section describes the framework used to evaluate the performance of the proposed NC-ALPHEE method in noise-free and noisy environments. The evaluation compares its estimation accuracy with that of the previously proposed ALPHEE method in Vimalajeewa et al. (2025) and the standard wavelet spectrum approach described in section 2.2.1. Simulated signals with Hurst exponent values ranging from 0.1 to 0.8 ($H \in [0.1, 0.8]$) were analyzed under varying levels of Gaussian noise ($\sigma_\epsilon \in [0, 1]$) to assess the robustness of each method under noise-free ($\sigma_\epsilon = 0$) and noisy ($\sigma_\epsilon > 0$) conditions.

### 4.1  Data Simulation

Self-similar signals were simulated using a fractional Brownian motion (fBm) process of length $N = 2^{16}$ with predefined Hurst exponent values. Additive Gaussian noise with zero mean and known variance $\sigma_\epsilon^2$ (i.e., $\epsilon \sim \mathcal{N}(0, \sigma_\epsilon^2)$) was added to emulate noisy environments. The noisy signal was then decomposed using the Discrete Wavelet Transform (DWT) with a *Symmlet* filter of size 6 ($M = 3$ vanishing moments). The filter size satisfies the wavelet decorrelation property described in equation (7). At each decomposition level $j$ ($1 \leq j \leq 15$), the signal energy was calculated as the average squared wavelet coefficient: $S(j) = \overline{d_j^2}$, as defined in Section 2.2.1. This yielded 105 pairwise energy values ($\binom{15}{2} = 105$) used for estimating $H$.

The simulation was repeated for Hurst exponent values $H = 0.1, 0.2, \ldots, 0.8$ across different noise levels: $\sigma_\epsilon = 0, 0.1, 0.25, 0.5, 0.75, 1$.

## 4.2 Estimation of $H$

The Hurst exponent was estimated using three methods: the standard wavelet spectrum, ALPHEE, and NC-ALPHEE. Each method used slightly different sets of level-pairs. The standard wavelet spectrum and the ALPHEE/NC-ALPHEE methods (using traditional averaging) used level pairs from scales $j = 3$ to $j = 13$ (55 pairs in total), whereas the neural network-based ALPHEE/NC-ALPHEE used level pairs from $j = 3$ to $j = 15$ (78 pairs in total).

Levels $j = 1$ and $j = 2$ were excluded because equation (31) requires more than four coefficients ($n_j > 4$) for valid variance estimation. Levels $j = 14$ and $j = 15$ were excluded in traditional methods as they tend to capture mostly noise and outliers Katul et al. (2006). This exclusion is standard in the literature Vimalajeewa, Bruce & Vidakovic (2023), Vimalajeewa et al. (2025), ensuring a fair comparison among methods. Consequently, 55 level pairs were used for the standard and ALPHEE methods, while 78 level pairs were used for the NN-based ALPHEE/NC-ALPHEE method.

### 4.2.1 Standard Wavelet Spectrum Method

As described in Section 2.2.1, this method estimates $H$ by fitting a linear regression model to the log-energies across scales $j = 3$ to $j = 13$. Estimation was repeated 1,000 times for each $H \in \{0.1, 0.2, \ldots, 0.8\}$ under each noise level.

### 4.2.2 ALPHEE and NC-ALPHEE Methods

Estimates of $H$ were computed for each simulated signal using Equations (17) and (14), resulting in multiple candidate estimates per signal. These were then aggregated to compute the final estimate using two different strategies:

#### 4.2.2.1 Traditional Averaging Approach:

From the 55 level pairs ($j = 3$ to $j = 13$), candidate estimates were aggregated using both arithmetic and weighted means, as well as the median. Weights for the weighted estimators were computed as the inverse variances of the corresponding estimates (see Section 3.5). As with the standard method, this process was repeated 1,000 times per $H$ value.

#### 4.2.2.2 Neural Network-Based Approach:

For each signal, 78 candidate estimates (from level pairs $j = 3$ to $j = 15$) were generated, resulting in an $8000 \times 78$ matrix (1,000 sample estimate sets per $H$ value obtained from repeated estimations). A fully connected feedforward neural network (NN) was trained to map these features to the corresponding Hurst exponent values.

The NN's hyperparameters were optimized using the Optuna framework. Tunable parameters included the number of hidden layers (2–5), neurons per layer (4–512), activation function (ReLU, Leaky ReLU, Tanh), learning rate, batch size, and weight decay. Leaky ReLU was often selected for improved gradient flow. The output layer consisted of a single neuron with linear activation for regression.

The dataset was standardized and randomly shuffled. It was split into training (85%) and testing (15%) sets. The training set was further divided using 5-fold cross-validation. The model was trained using mini-batch gradient descent and the Adam optimizer for up to 100 epochs, using mean squared error (MSE) as the loss function. Early stopping with a patience of 5 epochs and Optuna's pruning were employed to prevent overfitting. The best model was selected based on the lowest average validation loss.

The NN-based method was evaluated at noise levels $\sigma_\epsilon = 0, 0.25, 0.5, 0.75, 1.0$.

## 4.3 Performance Comparison

The performance of all three methods was compared under both noise-free ($\sigma_\epsilon = 0$) and noisy conditions ($\sigma_\epsilon > 0$). Simulations were conducted with noise levels $\sigma_\epsilon \in \{0, 0.25, 0.50, 0.75, 1.00\}$, as defined in Equation (6). As described in the previous sections, the same experimental setups were used under the traditional averaging and NN methods. The analysis focused on evaluating the performance of the NC-ALPHEE method in noisy conditions, comparing it with traditional averaging and NN methods.

This experimental setup enabled a comprehensive assessment of the robustness and accuracy of the proposed NC-ALPHEE method in comparison to the standard and ALPHEE methods across a range of noise intensities.

# 5 Results

This section presents the performance of the NC-ALPHEE method in estimating $H$. Initially, results are presented separately for the weighted median and NN approaches under both noise-free and noisy conditions. Subsequently, we present outcomes of the comparison, incorporating other averaging methods such as the arithmetic mean and median, across both noise-free and noisy environmental settings.

## 5.1 Weighted Median-based Estimation of $H$

### 5.1.1 Noise-free Conditions ($\sigma_\epsilon = 0$)

Figure 1 presents a comparative analysis of three methods for estimating the Hurst exponent ($\hat{H}$) across a range of actual Hurst exponent values ($H$) from 0.1 to 0.8. Each group of boxplots corresponds to a different true $H$, showing the distribution of estimates obtained using the *Standard* method (blue), the *ALPHEE* method (orange), and the *NC-ALPHEE* method under noise-free conditions (i.e., $\sigma_\epsilon = 0$, yellow). The Standard method exhibits greater variability across all $H$ values. The ALPHEE and NC-ALPHEE methods consistently yield estimates with lower variance and tighter concentration around the true values, indicating higher precision and robustness. NC-ALPHEE estimates show a slight reduction in variance as compared to the ALPHEE method. Overall, this analysis highlights that the pairwise estimation methods are more effective, specifically for higher $H$ values (i.e., $H \geq 0.3$) than the standard method in accurately estimating the $H$ in the absence of noise.

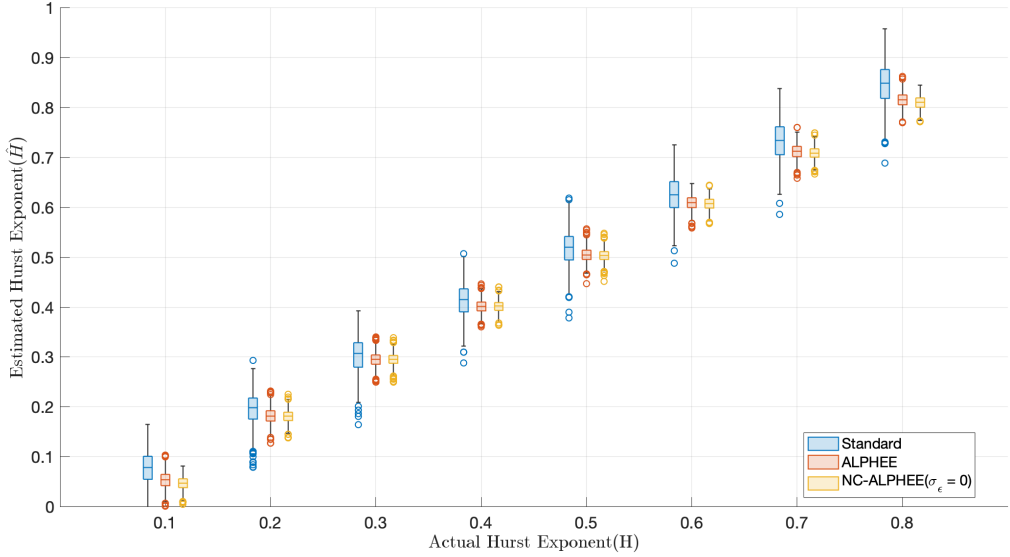

Figure 1: Comparison of Hurst exponent estimates obtained using the standard, ALPHEE, and NC-ALPHEE methods across varying actual Hurst exponent values under noise-free conditions ($\sigma_\epsilon = 0$, yellow).

### 5.1.2 Noisy Conditions ($\sigma_\epsilon > 0$)

Figure 2 shows the estimates of $H$ from the *Standard Wavelet Spectrum* (left), *ALPHEE* (middle), and *NC-ALPHEE* (right) under different levels of additive Gaussian noise ($\sigma_\epsilon = 0.10, 0.25, 0.50, 0.75$, and $1.00$). The level pairs considered in these estimations range from the scale 3 to 13 (i.e., $j_1 = 3, j_2 = 13$). The reason for using this scale range is to have a fair comparison among the three methods, though the NN-based approach is capable of using scales up to 15.

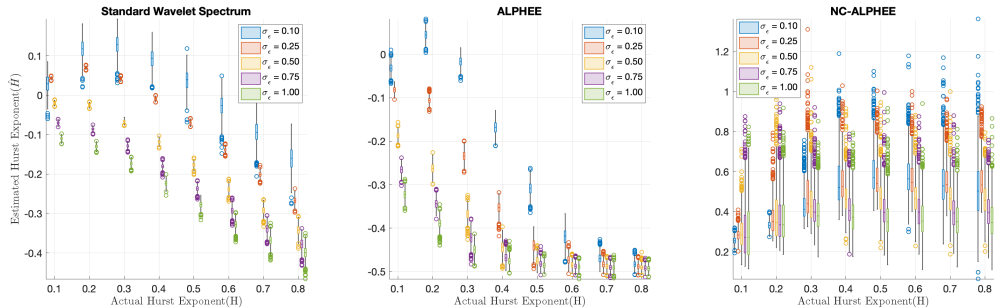

Figure 2: Comparison of three methods for estimating the Hurst exponent ($\hat{H}$) using all level-pairs between levels 3 and 13 and varying noise levels ($\sigma_\epsilon$).

Across all three methods, estimates show a significant underestimation as noise increases, affecting all $H$ values. More specifically, the estimates of the standard and ALPHEE methods deviate significantly from the true values, exhibiting strong underestimation. The NC-ALPHEE method demonstrates the best resilience to noise. Although the NC-ALPHEE still shows some underestimation as contamination increases, the effect is less severe compared to the other methods, which are giving negative estimates.

To investigate whether this underestimation is related to the set of level pairs used for estimation, under the same noisy conditions, we analyzed their performance by varying the range of levels considered. The outcomes from the NC-ALPHEE method are illustrated in Figure 3 for $H = 0.5$ (outcomes from the standard and ALPHEE are given in Figure 10 in Appendix). Each subplot represents a distinct range of scales used to generate level-pairs, starting from $[3, 7]$ and extending to $[3, 15]$, with boxplots showing the distribution of estimates for increasing noise levels ($\sigma_\epsilon = 0, 0.25, 0.50, 0.75, 1$). For example, the first subplot considered all possible level pairs within the levels 3 and 7. The dashed red line indicates the true Hurst exponent value ($H = 0.5$) used to generate the signal.

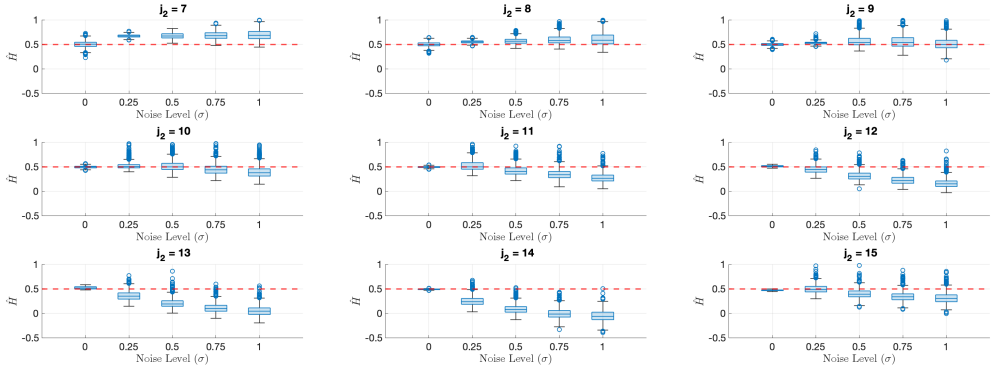

Figure 3: Analysis of the noise level ($\sigma_\epsilon$) and scale range on the estimation of the Hurst exponent ($\hat{H}$) using the NC-ALPHEE method. The true Hurst exponent ($H = 0.5$) is shown as a red dashed line, with the lower scale fixed at $j_1 = 3$ and the upper scale $j_2$ varying from 7 to 15.

When the scale range is narrow, which corresponds to the energies at coarser levels pairs (i.e., $j_1 = 3, j_2 \leq 8$), an increasing overestimation is visible as the noise increases. In contrast, an increasing underestimation is observed for wider scale ranges, which correspond to energies in both the coarser and finer level pairs (approximately $j_1 = 3, j_2 \geq 11$). As the permissible scale range increases (i.e., $j_1 = 3, 9 \leq j_2 \leq 11$), the variability decreases, and the bias in $\hat{H}$ reduces, resulting in estimates that cluster closer to the true $H$. In this range, the estimates become more stable and robust even under higher noise contamination, although a slight underestimation persists as $\sigma_\epsilon$ increases.

Figure 4, for instance, shows the NC-ALPHEE estimates for the level pairs used in the scale range from 3 to 10 (i.e., $j_1 = 3, j_2 = 10$) with increasing noise levels ($\sigma_\epsilon = 0, 0.25, 0.50, 0.75, 1$). For small values of $H$ (e.g., $H \leq 0.3$), the estimator remains robust across all noise levels. However, as the true $H$ increases, the estimator becomes increasingly biased and variable in the presence of noise. At higher $H$ values ($H \geq 0.5$), the estimates exhibit significant spread and frequent overestimation under moderate to high noise. The dashed red lines

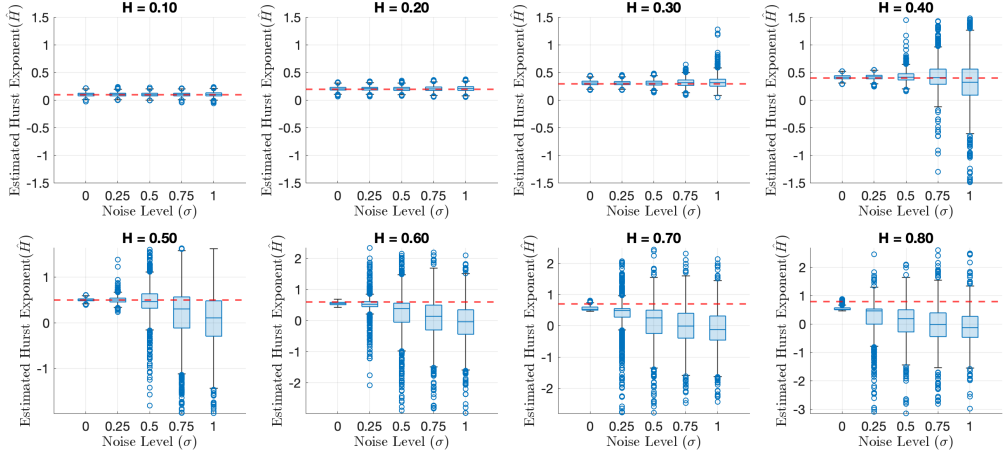

Figure 4: Effect of noise level ($\sigma_\epsilon$) on the estimation of the Hurst exponent ($\hat{H}$) for different true $H$ values using scale separation $|j_1 - j_2| = 3$ with scales ranging from 3 to 10.

represent the true $H$ values, helping visualize the deviation and variability introduced by noise.

This analysis underscores the critical role of selecting an optimal scale range in the estimation process, depending on the contamination level. While the standard and ALPHEE show poor estimation performance, the NC-ALPHEE can effectively mitigate the noise impact and accurately estimate $H$ when a suitable level range is selected. Excessively large ranges amplify noise sensitivity, leading to degraded accuracy, while overly narrow ranges reduce the number of level-pairwise energy values available for estimation, limiting performance. Mid-size scale ranges strike a balance, enhancing estimation accuracy by mitigating noise effects and ensuring sufficient data for robust calculations.

## 5.2 NN-based Estimation of $H$

### 5.2.1 Noise-free Conditions ($\sigma_\epsilon = 0$)

Following the same process used in Section 5.1.1, the NN model described in Section 3.5 was employed to compute the final estimate for $H$ under noise-free conditions. Table 1

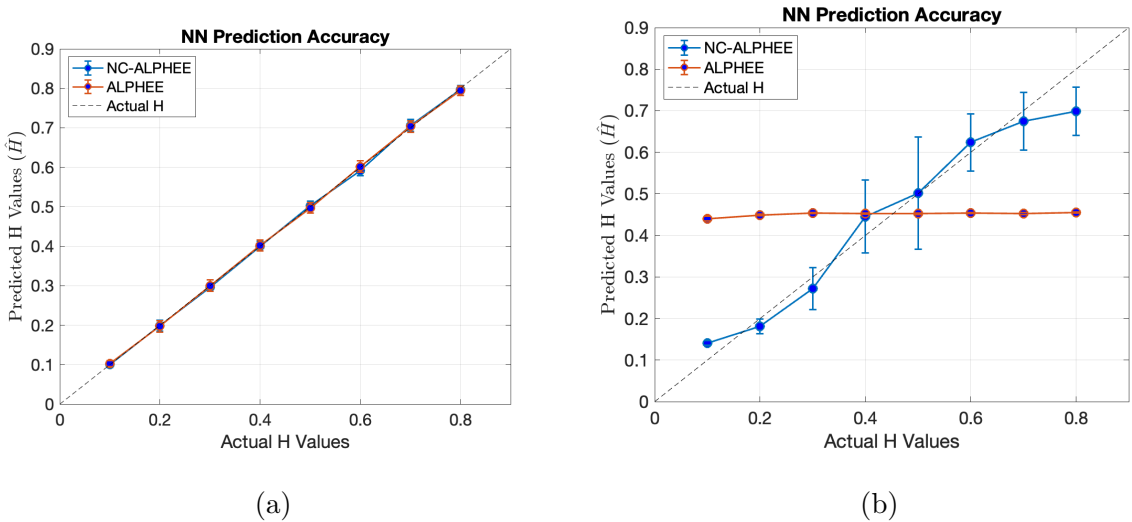

(a)            (b)

Figure 5: Hurst exponent estimations from each of the level-pairwise estimator integrated with neural network under (a) noise-free conditions ($\sigma_\epsilon = 0$) and (b) noisy conditions ($\sigma_\epsilon = 1.00$).

provides the NN configurations and Figure 5a shows the outcomes.

The dashed diagonal line represents the ideal case where predicted values perfectly match the actual ones. The solid orange line with error bars shows the mean predicted $\hat{H}$ values and their standard deviations (mean $\hat{H} \pm std(\hat{H})$) across multiple trials. The close alignment of the predicted values with the diagonal indicates that the NN approach achieves high accuracy in estimating $H$, with minimal bias and consistent performance across the tested range.

### 5.2.2 Noisy Conditions ($\sigma_\epsilon > 0$)

By running the same analysis described in the previous section with $\sigma_\epsilon = 1.00$, the outcomes are shown in Figure 5b. It is visible that the estimates from the NC-ALPHEE method are highly dense around their actual $H$ value, while an overestimation is indicated for the estimates from the ALPHEE method. Table 1 summarizes the NN settings used to produce this result, along with NN settings at other noise levels.

Table 1 summarizes the optimal neural network configurations for predicting $H$ across different noise levels ($\sigma_\epsilon$) in the data. For each noise setting, the table lists the number of units in each hidden layer (HL1–HL5), learning rate (LR), activation function (AF), batch size, weight decay, and the corresponding validation mean squared error (MSE). At low noise levels (0.00–0.50), the models achieved very low MSE values in the order of $10^{-5}$, indicating high predictive accuracy, whereas higher noise levels (0.75 and 1.50) led to notable increases in MSE, reflecting reduced performance under noisier conditions. The optimal architectures varied with noise level, with some configurations employing all five hidden layers and others omitting higher layers. Adjustments to learning rate, batch size, and weight decay across noise settings highlight the role of hyperparameter tuning in adapting model complexity and regularization to data quality.

| Noise Level ($\sigma_\epsilon$) | NN Configurations | | | | | | AF | Batch Size | Weight Decay | MSE |
| --- | --- | --- | --- | --- | --- | --- | --- | --- | --- | --- |
| | HL1 | HL2 | HL3 | HL4 | HL5 | LR | | | | |
| 0.00 | 484 | 68 | 228 | 324 | 324 | $5.8 \times 10^{-4}$ | Leaky ReLU | 64 | $9.18 \times 10^{-6}$ | $1.9 \times 10^{-5}$ |
| 0.25 | 132 | 260 | 68 | 4 | 164 | $9.5 \times 10^{-3}$ | tanh | 32 | $7.12 \times 10^{-6}$ | $2.0 \times 10^{-5}$ |
| 0.50 | 292 | 164 | 324 | 484 | – | $3.94 \times 10^{-4}$ | Leaky ReLU | 16 | $1.10 \times 10^{-6}$ | $2.9 \times 10^{-5}$ |
| 0.75 | 388 | 4 | 324 | 228 | 484 | $4.71 \times 10^{-4}$ | Leaky ReLU | 16 | $2.29 \times 10^{-6}$ | $7.73 \times 10^{-4}$ |
| 1.00 | 324 | 132 | 356 | 196 | – | $5.52 \times 10^{-4}$ | Leaky ReLU | 16 | $1.07 \times 10^{-6}$ | $2.811 \times 10^{-3}$ |

Table 1: Optimal NN model configurations under different noise levels and their $H$ prediction performance. HL = hidden layer, LR = learning rate, AF = activation function, MSE = validation mean squared error.

## 5.3 Performance Comparison

Figure 6 compares the predicted Hurst exponents ($\hat{H}$) against the true Hurst exponents ($H$) under three noise levels: $\sigma_\epsilon = 0.0, 0.5,$ and $1.0$. The orange dashed line represents the ideal case where predicted values perfectly match the true values. Under noise-free conditions ($\sigma_\epsilon = 0.0$), all aggregation methods closely follow the ideal line, with the neural network (NN) predictions aligning almost perfectly. As noise increases to $\sigma_\epsilon = 0.5$, the NN maintains high predictive accuracy, whereas the weighted and arithmetic aggregation methods show significant deviations. At the highest noise level ($\sigma_\epsilon = 1.0$), the NN continues to track the ideal line closely, highlighting its resilience to noise, while the other methods exhibit systematic underestimation of $H$. These results demonstrate that the NN-based aggregation approach provides superior accuracy and noise resilience compared to traditional aggregation methods.

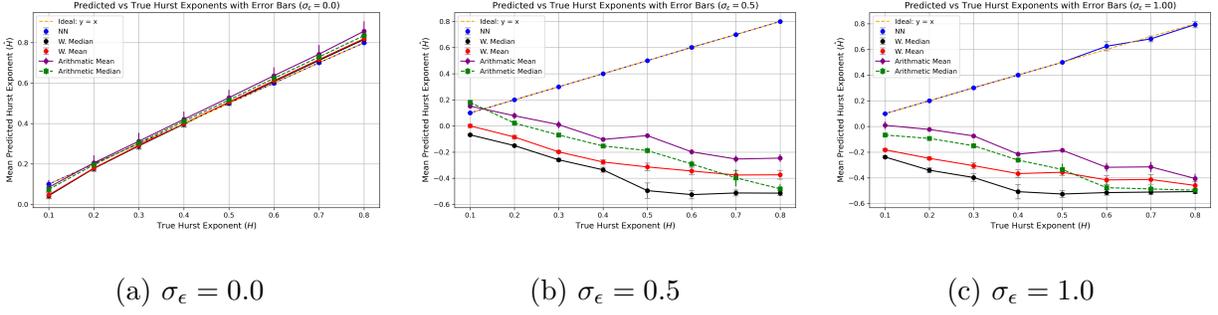

(a) $\sigma_\epsilon = 0.0$  (b) $\sigma_\epsilon = 0.5$  (c) $\sigma_\epsilon = 1.0$

Figure 6: Predicted Hurst exponents ($\hat{H}$) compared to true values ($H$) across five aggregation techniques—Neural Network (NN), Weighted Median, Weighted Mean, Arithmetic Mean, and Arithmetic Median—under three noise levels: (a) $\sigma_\epsilon = 0.0$, (b) $\sigma_\epsilon = 0.5$, and (c) $\sigma_\epsilon = 1.0$. Each point represents the mean of 1000 simulation trials for each true Hurst value ($H$), with error bars indicating standard deviation.

The three plots in Figure 7 illustrate the comparison between the true sample counts and the predicted sample counts across Hurst exponent bins under varying noise conditions ($\sigma_\epsilon = 0.0, 0.5$, and $1.0$). In the noise-free case ($\sigma_\epsilon = 0.0$), the predicted counts closely align with the true counts for all bins, demonstrating the neural network's ability to accurately aggregate candidate estimators when no external noise is present. As the noise level increases to $\sigma_\epsilon = 0.5$, the predictions remain robust, with only slight deviations observed in bins corresponding to higher Hurst exponents. At the highest noise level ($\sigma_\epsilon = 1.0$), the model exhibits a more noticeable divergence from the true counts, particularly at the extremes of the Hurst exponent range. This pattern suggests that while the neural network is highly effective in noise-free and moderately noisy environments, its performance is modestly impacted under heavy noise, reflecting the inherent difficulty of estimation under such conditions.

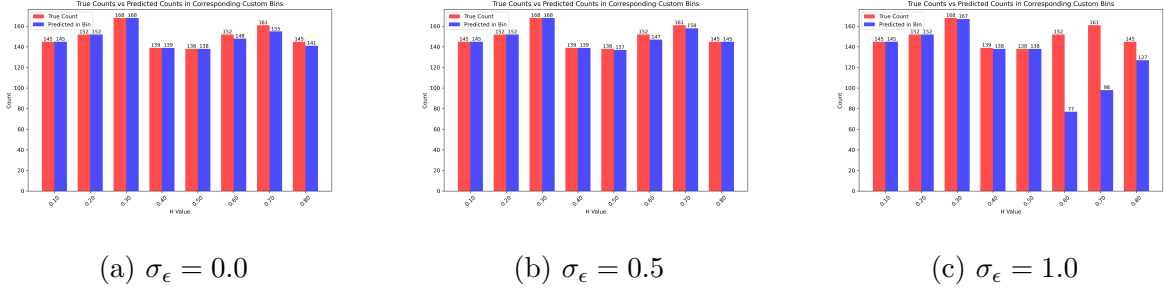

(a) $\sigma_\epsilon = 0.0$  (b) $\sigma_\epsilon = 0.5$  (c) $\sigma_\epsilon = 1.0$

Figure 7: Comparison of true and predicted sample counts within proposed Hurst exponent ($H$) bins under three noise levels:(a) $\sigma_\epsilon = 0.0$, (b) $\sigma_\epsilon = 0.5$, and (c) $\sigma_\epsilon = 1.0$. Each bar plot shows the number of samples falling into each $H$ bin for both the true counts (red) and the predicted counts (blue). The height of the bars, along with the counts labeled on top, highlights how prediction accuracy changes as noise increases, particularly showing a divergence in counts for higher noise levels.

# 6 Discussion

This section discusses the unique properties of the proposed NC-ALPHEE method and how they contribute to overcoming limitations of existing methods, supported by insights from simulation studies.

Self-similar signals typically exhibit a wavelet spectrum that follows a linear pattern, as described in equation (5). However, when such signals are contaminated by noise, the wavelet spectrum deviates from this linearity. Specifically, the spectrum shows a linear decay at coarser levels, followed by a relatively flat region at finer levels (see Figure 8). This deviation becomes more pronounced as the level of noise increases, a phenomenon commonly referred to as the *"hockey-stick effect"* in Katul et al. (2006). The region of linear decay corresponds to the inherent self-similarity of the signal, whereas the flat region reflects the influence of noise. This limits the standard wavelet spectrum- and ALPHEE-based $H$ estimation, particularly in noisy conditions. This is because, in finer levels, the slope is close to zero due to the fact that the small log energy difference, $\log_2 \overline{d_{j_1}^2} - \log_2 \overline{d_{j_2}^2} \approx 0$. Consequently, this limits the range of levels to be used to fit a regression line in the standard method. In the ALPHEE method given in (17), this energy difference is represented as the bias correction term and hence the estimation relies mainly on the difference between the digamma functions $\psi(n_{j_1}/2) - \psi(n_{j_2}/2)$. This is why these existing methods show poor performance in noisy conditions. Whereas, the NC-ALPHEE in equation (14) updates this bias correction by subtracting the noise influence as $n_{j_i} \overline{d_{j_1}^2} - 2\sigma_\epsilon^2 \exp \psi(n_{j_i}/2)$. This, in turn, helps consider level-pairwise energies even from finer levels for the estimation.

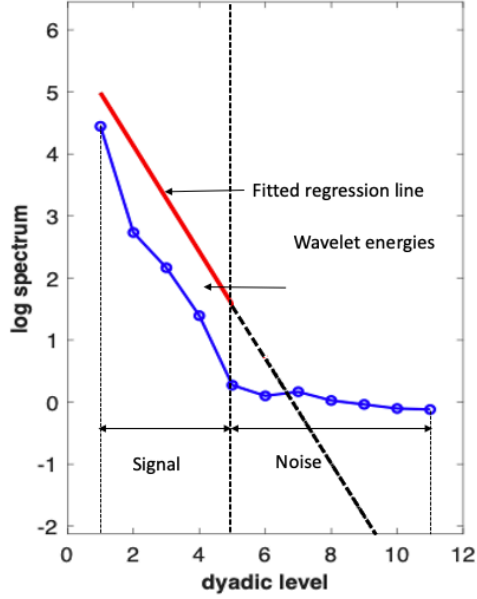

Figure 8: The hockey stick nature of wavelet spectrum of a contaminated signal. A linear decay in wavelet energy (y-axis) at lower dyadic (coarser) levels, followed by a relatively flat region towards the higher dyadic (finer) levels.

The analysis also underscores the need for an effective aggregation strategy for pairwise level-based methods to ensure accurate $H$ estimation in noisy signals. The traditional averaging-based aggregation methods exhibit reduced accuracy in estimating $H$ under conditions of significant signal contamination. As highlighted in Vimalajeewa et al. (2025), higher levels of noise can diminish the availability of reliable level pairs, thereby limiting the effective utilization of level-pairwise energy information. For instance, as shown in Figure 1 and 6a, for noise-free signals, aggregating candidate estimates from both ALPHEE and NC-ALPHEE using the traditional averaging methods produces reliable estimates of $H$. However, Figure 2, 6b, and 6c show that this approach is less effective for noisy signals. In particular, computing weights by inverting the variance, as defined in Equation (19), does not provide optimal results under high noise contamination.

To mitigate the limitations of existing aggregation techniques, we use an alternative approach leveraging a neural network, which is designed to learn from the available estimates and enhance the robustness of $H$ estimation in noisy environments. Figure 6b and Figure 6c demonstrate that employing a neural network (NN) to aggregate candidate estimates from NC-ALPHEE achieves accurate $H$ estimation even in the presence of significant noise. Unlike traditional methods, which rely on fixed, simple rules, NNs help improve aggregation of candidate estimators because they learn complex, adaptive, and nonlinear combinations of the candidates. In addition, NNs automatically discover how to weight each estimator based on the specific input context, effectively handle correlated errors between estimators, and leverage subtle patterns in the data. This allows them to model intricate relationships and adjust the fusion strategy for higher accuracy, where traditional techniques fall short. Overall, these findings indicate that integrating NC-ALPHEE with an NN provides a robust and adaptive solution for estimating the Hurst exponent in noisy environments.

It is also worth noting that the proposed method still poses some challenges, which can be seen under theoretical and practical aspects. Under the theoretical aspects, this study primarily focused on a single $H$ value to describe global self-similarity. In many practical settings (e.g., financial time series), the $H$ values could be varied over time, so that $H$ has to be modeled as a function of time in order to characterize self-similarity more precisely. A potential approach is to explore how the proposed method can be updated by using the Hurst exponent estimation of locally self-similar processes method discussed in Coeurjolly (2008).

Considering the practical aspects, optimizing NN settings for longer time series is resource-intensive, especially with the increasing number of candidate estimates. While NC-ALPHEE accounts for all possible level pairs, not all contribute to estimating $H$. To improve estimation efficacy, the NN structure can be enhanced, for example, by incorporating dropout layers. Additionally, utilizing the variances of candidate estimates to initialize NN weights can streamline training, reducing computational costs and time—critical for large-scale datasets.

To further evaluate the reliability of the pairwise Hurst exponent estimates, we examined their behavior within the uncertainty bounds of the aggregated predictions. As shown in Fig. **??**(a), for each true $H$ value we first determined the $2\sigma$ band around the mean of the aggregated predictions. We then filtered all instances where the predicted values fell within this band, ensuring that the analysis focused on the range capturing the majority of plausible predictions. Next, we mapped the 78 pairwise estimates to their corresponding $(j_1, j_2)$ scale pairs $(j_1 < j_2)$ and counted how often each pair produced an estimate within the defined band. The resulting triangular heatmap in Fig. **??**(b) provides a visual representation of the stability of different scale pairs. Brighter cells indicate pairs that consistently yield reliable estimates, whereas darker regions highlight pairs that rarely contribute valid information. Overall, the heatmap reveals that certain low-to-mid scale combinations (e.g., $(j_1, j_2)$ near the lower-left region) are particularly stable, while higher-scale pairs tend to produce noisier

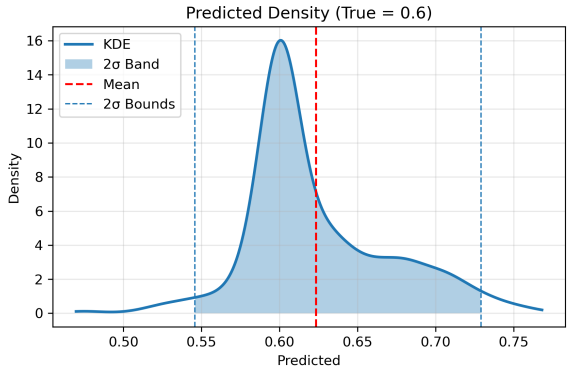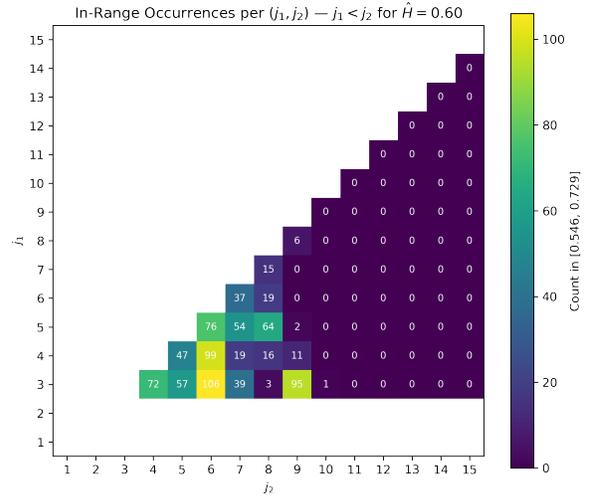

(a) Kernel density of predicted $H$ for true $H = 0.6$.

(b) Heatmap of in-range occurrences for true $H = 0.6$.

Figure 9: Comparison of predicted $H$ distributions and pairwise estimate reliability for true $H = 0.6$. Panel (a) shows the kernel density of aggregated predictions with mean and $\pm 2\sigma$ bounds. Panel (b) highlights the pairwise scale combinations $(j_1, j_2)$ contributing estimates within the $2\sigma$ band.

estimates.

# 7 Conclusion

This study presents a novel method, NC-ALPHEE, for estimating the Hurst exponent $H$ in noisy signals. The method is derived by analyzing the exact expectations and variances of wavelet coefficients, taking into account their distributional characteristics and decorrelation behavior. NC-ALPHEE generates a set of candidate $H$ estimates by examining pairwise signal energies across multiple resolution levels obtained through wavelet decomposition. The previously proposed ALPHEE method is a special case of NC-ALPHEE, applicable in noise-free settings.

Simulation results demonstrate that NC-ALPHEE performs as well as or better than existing methods. In noise-free environments, it shows comparable accuracy to both the standard wavelet spectrum and ALPHEE methods when using either traditional averaging or neural network (NN)-based aggregation techniques. In noisy conditions, the averaging-based NC-ALPHEE remains effective by selectively restricting the level pairs used in estimation—though this selection becomes more difficult when noise levels vary across signals. In contrast, the NN-based version consistently delivers strong performance across all noise levels without the need to limit level pairs.

NC-ALPHEE provides a robust and noise-tolerant framework for self-similarity analysis, excelling in noisy conditions where conventional methods often fall short. By integrating a neural network (NN) for aggregating candidate estimates, the method achieves more reliable and adaptive results compared to traditional averaging techniques.

In the spirit of reproducible research, the software used in this paper is available at *[GitHub](GitHub)* repository.

# 8 Disclosure statement

The authors have no conflicts of interest.

# 9 Data Availability Statement

The software used for the simulations has been made available at the URL *[GitHub](GitHub)* repository

**SUPPLEMENTARY MATERIAL**

**Estimation of $\hat{H}$**

In order to find an estimator for $H$, we take into account the relationship shown in equation (12) at two different levels. To do that, we replace $n_j$ and $j$ in equation (12) with $n_{j_i}$ and $j_i$ as:

$$\log n_{j_i} + \mathbb{E}\left(\log(\overline{d_{j_i}^2})\right) - \log\left(\sigma_X^2 2^{-j_i(2H+1)} + \sigma_\epsilon^2\right) = \log 2 + \psi\left(\frac{n_{j_i}}{2}\right), \quad i = 1, 2, \quad (22)$$

where $n_{j_i} = 2^{j_i}$ is the number of wavelet coefficients at level $j_i$.

Consider the relationship in equation (22) at two different levels $j_1$ and $j_2$.

$$\log n_{j_1} + \mathbb{E}\left(\log(\overline{d_{j_1}^2})\right) - \log\left(\sigma_X^2 2^{-j_1(2H+1)} + \sigma_\epsilon^2\right) = \log 2 + \psi\left(\frac{n_{j_1}}{2}\right) \quad (23)$$

$$\log n_{j_2} + \mathbb{E}\left(\log(\overline{d_2^2})\right) - \log\left(\sigma_X^2 2^{-j_2(2H+1)} + \sigma_\epsilon^2\right) = \log 2 + \psi\left(\frac{n_{j_2}}{2}\right) \quad (24)$$

In the equation (23) and (24), using the method of moments, the first moment of the population $\mathbb{E}(\log(\overline{d_{j_i}^2}))$ is replaced by the first moment of the sample of the wavelet coefficients $\log \overline{d_{j_i}^2}$. The resulting expressions are then rearranged as follows.

$$-j_1(2H+1) = \log_2\left(\frac{n_{j_1}\overline{d_{j_1}^2} - 2\sigma_\epsilon^2 e^{\psi(n_{j_1}/2)}}{2\sigma_X^2 e^{\psi(n_{j_1}/2)}}\right). \quad (25)$$

$$-j_2(2H+1) = \log_2\left(\frac{n_{j_2}\overline{d_{j_2}^2} - 2\sigma_\epsilon^2 e^{\psi(n_{j_2}/2)}}{2\sigma_X^2 e^{\psi(n_{j_2}/2)}}\right). \quad (26)$$

The difference between equation (25) and (26) removes the unknown parameter $\sigma_X^2$.

$$(j_1 - j_2)(2H + 1) = \log_2\left(\frac{(n_{j_1}\overline{d_{j_1}^2} - 2\sigma_\epsilon^2 e^{\psi(n_{j_1}/2)})}{(n_{j_2}\overline{d_{j_2}^2} - 2\sigma_\epsilon^2 e^{\psi(n_{j_2}/2)})} \frac{e^{\psi(n_{j_1}/2)}}{e^{\psi(n_{j_2}/2)}}\right). \quad (27)$$

This can be simplified to obtain the expression for estimating $H$ as follows

$$\hat{H} = \frac{1}{2(j_1 - j_2)} \left[ \left( \frac{\psi(n_{j_1}/2) - \psi(n_{j_2}/2)}{\log(2)} \right) - \log_2 \left( \frac{n_{j_2}\overline{d_{j_2}^2} - 2\sigma_\epsilon^2 e^{\psi(n_{j_2}/2)}}{n_{j_1}\overline{d_{j_1}^2} - 2\sigma_\epsilon^2 e^{\psi(n_{j_1}/2)}} \right) \right] - \frac{1}{2}, \quad (28)$$

where $0 \leq j_1 < j_2 \leq J - 1$ and $J = \log_2 N$, with $N$ being the length of the signal.

## Variance of $\hat{H}$

According to the properties of the well-known log-chi-squared distribution, if $W \sim \log \chi_n^2$, then variance of $W$ are given

$$Var(W) = \psi'(n/2), \quad (29)$$

where $\psi'$ is trigamma function.

Using these two expressions (29) (28), the variance of $\hat{H}$ can be expressed as follows:

$$V(\hat{H}) = \frac{1}{4(j_1 - j_2)^2} \left[ V\left( \log_2(n_{j_1}\overline{d_{j_1}^2} - 2\sigma_\epsilon^2 e^{\psi(n_{j_1}/2)}) \right) + V\left( \log_2(n_{j_2}\overline{d_{j_2}^2} - 2\sigma_\epsilon^2 e^{\psi(n_{j_2}/2)}) \right) \right], \quad (30)$$

where $0 \leq j_1 < j_2 \leq J - 1$ and $J = \log_2 N$, with $N$ being the length of the signal.

Let $M_i = \log_2(n_{j_i}\overline{d_{j_i}^2} - 2\sigma_\epsilon^2 e^{\psi(n_{j_i}/2)})$ for $i = 1, 2$, and express $V(\hat{H})$ in equation (30) in a more abstract form as

$$V(\hat{H}) = \frac{1}{4(j_1 - j_2)^2} \left[ V(M_1) + V(M_2) \right] \quad (31)$$

Assume that the bias correction term $2\sigma_\epsilon^2 e^{\psi(n_{j_i}/2)}$ in $M_i$ is a small perturbation compared to $n_{j_i}\overline{d_{j_i}^2}$. Using the Taylor expansion, the first-order approximation of $M_i$ is

$$M_i = \log_2(n\overline{d_{j_i}^2}) - \frac{2\sigma_\epsilon^2 e^{\psi(n_{j_i}/2)}}{\log(2)n_{j_i}} \frac{1}{\overline{d_{j_i}^2}} + O(\sigma_\epsilon^4) \approx \log_2(n\overline{d_{j_i}^2}) - \frac{2\sigma_\epsilon^2 e^{\psi(n_{j_i}/2)}}{\log(2)n_{j_i}} \frac{1}{\overline{d_{j_i}^2}} \quad (32)$$

Variance of this first-order approximation of $M_i$ can be calculated as

$$V(M_i) = V(A) + V(B) + 2Cov(A,B) \quad \text{where} \quad A = \log_2(n_{j_i}\overline{d_{j_i}^2}) \quad \text{and} \quad B = -\frac{2\sigma_\epsilon^2 e^{\psi(n_{j_i}/2)}}{\log(2)n_{j_i}} \frac{1}{\overline{d_{j_i}^2}} \tag{33}$$

In the following, we compute each variance component in equation (33) using the given assumptions and known moments of $\overline{d_{j_i}^2} \sim \chi_{n_{j_i}}^2$.

1. **Variance of $A = \log_2(n_{j_i}\overline{d_{j_i}^2})$**: Given the log-chi-squared distribution properties of A via equation (10), variance of A variance is calculated as follows.

$$A = \log_2(n_{j_i}\overline{d_{j_i}^2}) = \frac{1}{\log(2)} \log(\overline{d_{j_i}^2}) \Rightarrow V(A) = \frac{1}{\log(2)^2} \psi'(n_{j_i}/2), \tag{34}$$

where $\psi'(n_{j_i}/2)$ is the trigamma function.

2. **Variance of $B = -\frac{2\sigma_\epsilon^2 e^{\psi(n_{j_i}/2)}}{\log(2)n_{j_i}} \frac{1}{\overline{d_{j_i}^2}}$**: Let $C = -\frac{2\sigma_\epsilon^2 e^{\psi(n_{j_i}/2)}}{\log(2)n_{j_i}}$ and then the variance of $B$ is

$$V(B) = C^2 V\left(\frac{1}{\overline{d_{j_i}^2}}\right) = C^2 \left(\mathbb{E}\left[\left(\frac{1}{\overline{d_{j_i}^2}}\right)^2\right] - \left[\mathbb{E}\left(\frac{1}{\overline{d_{j_i}^2}}\right)\right]^2\right) \tag{35}$$

Given that $L = \frac{n_{j_i}\overline{d_{j_i}^2}}{\sigma_{j_i}^2} \sim \chi_{n_j}^2$ with $\sigma_{j_i}^2$ being the variance of wavelet coefficients at the level $j_i$, $\frac{1}{\overline{d_{j_i}^2}} = \frac{n_{j_i}}{\sigma_{j_i}^2} \frac{1}{L}$. Using the moments of $\chi_n^2$ distribution,

$$\mathbb{E}\left(\frac{1}{\overline{d_{j_i}^2}}\right) = \frac{n_{j_i}}{\sigma_{j_i}^2} \mathbb{E}\left(\frac{1}{L}\right). \tag{36}$$

Recall, if a random variable $C$ has a $\chi_n^2$ distribution, then $1/C$ has an inverse chi-square distribution. This follows

$$C \sim \chi_n^2 \iff C \sim \text{Gamma}\left(\alpha = \tfrac{n}{2},\ \theta = 2\right) \implies \mathbb{E}\left[\frac{1}{C}\right] = \frac{1}{(\alpha-1)\theta},\ \alpha > 1,$$

.

Plugging $\alpha = n/2$ and $\theta = 2$ in $\mathbb{E}\left[\frac{1}{C}\right]$ gives

$$\mathbb{E}\left[\frac{1}{C}\right] = \frac{1}{n-2}, \qquad \text{for } n > 2.$$

Therefore, $\mathbb{E}\left(\frac{1}{L}\right) = \frac{1}{n_j - 2}$. Substituting this in equation (36),

$$\mathbb{E}\left(\frac{1}{\overline{d_{j_i}^2}}\right) = \frac{n_{j_i}}{\sigma_{j_i}^2}\mathbb{E}\left(\frac{1}{L}\right) = \frac{n_{j_i}}{\sigma_{j_i}^2}\frac{1}{(n_{j_i} - 2)} \quad \text{for} \quad n_{j_i} > 2. \tag{37}$$

Similarly,

$$\mathbb{E}\left[\left(\frac{1}{\overline{d_{j_i}^2}}\right)^2\right] = \frac{n_{j_i}^2}{\sigma_{j_i}^4}\frac{1}{(n_{j_i} - 2)(n_{j_i} - 4)}, \quad \text{for} \quad n_{j_i} > 4 \tag{38}$$

The substitution of expression for $\mathbb{E}\left(\frac{1}{\overline{d_{j_i}^2}}\right)$ and $\mathbb{E}\left[\left(\frac{1}{\overline{d_{j_i}^2}}\right)^2\right]$ given in equation (37) and (38) in equation (35) results in

$$V(B) = \frac{8\sigma_\epsilon^4 e^{2\psi(n_{j_i}/2)}}{\log(2)^2 \sigma_{j_i}^4 (n_{j_i} - 2)^2 (n_{j_i} - 4)}, \quad \text{for} \quad n_{j_i} > 4 \tag{39}$$

3. **Covariance of A and B, $Cov(A, B)$:**

$$\begin{aligned}
Cov(A, B) &= Cov\left(\frac{\log(\overline{d_{j_i}^2})}{\log(2)}, \frac{-2\sigma_\epsilon^2 e^{\psi(n_{j_i}/2)}}{\log(2) n_{j_i} \overline{d_{j_i}^2}}\right) \\
&= \frac{-2\sigma_\epsilon^2 e^{\psi(n_{j_i}/2)}}{\log(2)^2 n_{j_i}}\left[\mathbb{E}\left(\log(\overline{d_{j_i}^2})\frac{1}{\overline{d_{j_i}^2}}\right) - \mathbb{E}(\log \overline{d_{j_i}^2})\mathbb{E}\left(\frac{1}{\overline{d_{j_i}^2}}\right)\right]
\end{aligned} \tag{40}$$

Now, compute the two expectations in equation (40)) separately

- $\mathbb{E}\left(\log(\overline{d_{j_i}^2})\frac{1}{\overline{d_{j_i}^2}}\right)$: We can calculate this expectation by using the expression for $\log(L)$ and $L$ as follows:

$$\log(\overline{d_{j_i}^2}) = \log\left(\frac{\sigma_{j_i}^2}{n_{j_i}}\right) + \log(L)$$

$$\frac{\log(\overline{d_{j_i}^2})}{\overline{d_{j_i}^2}} = \frac{\log\left(\frac{\sigma_{j_i}^2}{n_{j_i}}\right) + \log(L)}{(\sigma_{j_i}^2/n_{j_i})L}$$

$$= \frac{n_{j_i}}{\sigma_{j_i}^2}\left[\log\left(\frac{\sigma_{j_i}^2}{n_{j_i}}\right)\frac{1}{L} + \frac{\log(L)}{L}\right]$$

$$\mathbb{E}\left(\frac{\log(\overline{d_{j_i}^2})}{\overline{d_{j_i}^2}}\right) = \frac{n_{j_i}}{\sigma_{j_i}^2}\left[\log\left(\frac{\sigma_{j_i}^2}{n_{j_i}}\right)\mathbb{E}\left(\frac{1}{L}\right) + \mathbb{E}\left(\frac{\log(L)}{L}\right)\right]. \tag{41}$$

In expression (37), we have already computed $\mathbb{E}\left(\frac{1}{L}\right) = \frac{1}{(n_{j_i}-2)}$ using the moments of $\chi_n^2$ distribution. Following the same method, we can compute $\mathbb{E}\left(\frac{\log(L)}{L}\right)$.

$$\mathbb{E}\left(\frac{\log(L)}{L}\right) = \int_0^\infty \left(\frac{\log(L)}{L}\right)f(l)dl, \quad \text{with} \quad f(l) = \frac{1}{2^{n_{j_i}}\Gamma(n_{j_i}/2)}l^{(n_{j_i}/2-1)}e^{-l/2}$$

By substituting $t = l/2$, the above integral can be simplified as follows

$$\mathbb{E}\left(\frac{\log(L)}{L}\right) = \frac{\log(2)}{\Gamma(n_{j_i}/2)}\int_0^\infty t^{(n_{j_i}/2-2)}e^{-t}dt + \frac{1}{\Gamma(n_{j_i}/2)}\int_0^\infty \log(t)t^{(t/2-2)}e^{-t}dt$$

$$= \frac{\log(2)}{\Gamma(n_{j_i}/2)}\Gamma(n_{j_i}/2 - 1) + \frac{\Gamma(n_{j_i}-1)\psi(n_{j_i}/2-1)}{\Gamma(n_{j_i}/2)}$$

$$= \left(\log(2) + \psi(n_{j_i}/2 - 1)\right)\frac{1}{(n_{j_i} - 2)}. \tag{42}$$

By substituting the expression for $\mathbb{E}\left(\frac{1}{L}\right)$ in equation (37) and $\mathbb{E}\left(\frac{\log(L)}{L}\right)$ in equation (41) in equation (42),

$$\mathbb{E}\left(\frac{\log(\overline{d_{j_i}^2})}{\overline{d_{j_i}^2}}\right) = \frac{n}{(n_{j_i} - 2)\sigma_{j_i}^2}\left[\log\left(\frac{2\sigma_{j_i}^2}{n_{j_i}}\right) + \psi\left(\frac{n_{j_i}}{2} - 1\right)\right] \tag{43}$$

- $\mathbb{E}(\log \overline{d_{j_i}^2})$ **and** $\mathbb{E}\left(\frac{1}{\overline{d_{j_i}^2}}\right)$:. According to the equation (38), $\mathbb{E}\left(\frac{1}{\overline{d_{j_i}^2}}\right) = \frac{n_{j_i}}{\sigma_{j_i}^2 (n_{j_i} - 2)}$.

  Based on the expression in (11), $\mathbb{E}[\log(\overline{d_{j_i}^2})]$ can be computed as follows:

$$\mathbb{E}\left[\log\left(\frac{n_{j_i}\overline{d_{j_i}^2}}{\sigma_{j_i}}\right)\right] = \psi(n_{j_i}/2) + \log(2)$$

$$\mathbb{E}[\log(\overline{d_{j_i}^2})] = \psi(n_{j_i}/2) + \log\left(\frac{2\sigma_{j_i}^2}{n_{j_i}}\right). \tag{44}$$

Combining the expressions given in equations (38) and (44), the second expectation term in equation (40) can be expressed as:

$$\mathbb{E}(\log \overline{d_{j_i}^2})\mathbb{E}\left(\frac{1}{\overline{d_{j_i}^2}}\right) = \frac{n_{j_i}}{\sigma_{j_i}^2(n_{j_i}-2)}\left(\psi\left(\frac{n_{j_i}}{2}\right) + \log\left(\frac{2\sigma_{j_i}^2}{n_{j_i}}\right)\right) \tag{45}$$

By substituting expressions in (43) and (45) in the equation (40), the covariance between $A$ and $B$ becomes

$$\begin{aligned} Cov(A,B) &= \frac{-2\sigma_\epsilon^2 e^{\psi(n_{j_i}/2)}}{\log(2)^2 n_{j_i}}\left[\frac{n_{j_i}}{\sigma_{n_{j_i}}^2(n_{j_i}-2)}\left(\log\left(\frac{2\sigma_{j_i}^2}{n_{j_i}}\right) + \psi\left(\frac{n_{j_i}}{2}-1\right)\right)\right. \\ &\qquad \left. - \frac{n_{j_i}}{\sigma_{j_i}^2(n_{j_i}-2)}\left(\psi\left(\frac{n_{j_i}}{2}\right) + \log\left(\frac{2\sigma_{j_i}^2}{n_{j_i}}\right)\right)\right] \\ &= \frac{\sigma_\epsilon^2 e^{\psi(n_{j_i}/2)}}{\log(2)^2 \sigma_{j_i}^2 (n_{j_i}-2)}\left(\psi(n_{j_i}/2-1) - \psi(n_{j_i}/2)\right) \\ &= \frac{4\sigma_\epsilon^2 e^{\psi(n_{j_i}/2)}}{\log(2)^2 \sigma_{j_i}^2 (n_{j_i}-2)n_{j_i}} \end{aligned} \tag{46}$$

Substituting $V(A), V(B)$, and $Cov(A, B)$ given respectively in equation (34), (39), and (46) in (33), we can compute $V(M_i)$ as follows.

$$V(M_i) \approx \frac{1}{\log(2)^2}\left[\psi'\left(\frac{n_{j_i}}{2}\right) + \frac{8\sigma_\epsilon^4 e^{2\psi(n_{j_i}/2)}}{\sigma_{j_i}^4(n_{j_i}-2)^2(n_{j_i}-4)} + \frac{8\sigma_\epsilon^2 e^{\psi(n_{j_i}/2)}}{\sigma_{j_i}^2(n_{j_i}-2)n_{j_i}}\right] \tag{47}$$

Finally, using the expression for $V(M_1)$ and $V(M_2)$ given in equation (47) into (31), we can compute the variance of $\hat{H}$

$$V(\hat{H}) \approx \frac{1}{\left(2(j_1 - j_2)\log(2)\right)^2} \sum_{i=1}^{2} \left[ \psi'\left(\frac{n_{j_i}}{2}\right) + \frac{8\sigma_\epsilon^4 e^{2\psi(n_{j_i}/2)}}{\sigma_{j_i}^4 (n_{j_i} - 2)^2 (n_{j_i} - 4)} + \frac{8\sigma_\epsilon^2 e^{\psi(n_{j_i}/2)}}{\sigma_{j_i}^2 (n_{j_i} - 2) n_{j_i}} \right] \quad (48)$$

To estimate $\hat{H}$ and $V(\hat{H})$, we need the noise variance $\sigma_\epsilon^2$ and variance of detail wavelet coefficients at decomposition level $j_i$, $\sigma_{j_i}^2$. According to Vidakovic (1999), the variance of the finest level (i.e., $j = J - 1$) of detail wavelet coefficients is a good estimator of the noise variance.

$$\sigma_\epsilon^2 \approx V(d_{jk}), \quad \text{where} \quad j = J - 1$$

The $\sigma_{j_i}^2$ is estimated as the average square of the detail wavelet coefficients at the level $j_i$th level.

$$\sigma_{j_i}^2 = V(d_{j_i}) = \mathbb{E}(d_{j_i}^2) - [\mathbb{E}(d_{j_i})]^2 = \mathbb{E}(d_{j_i}^2) \approx \overline{d_{j_i}^2} \quad (\because \mathbb{E}(d_{j_i}) = 0)$$

**Scale Range in Noisy Signals**

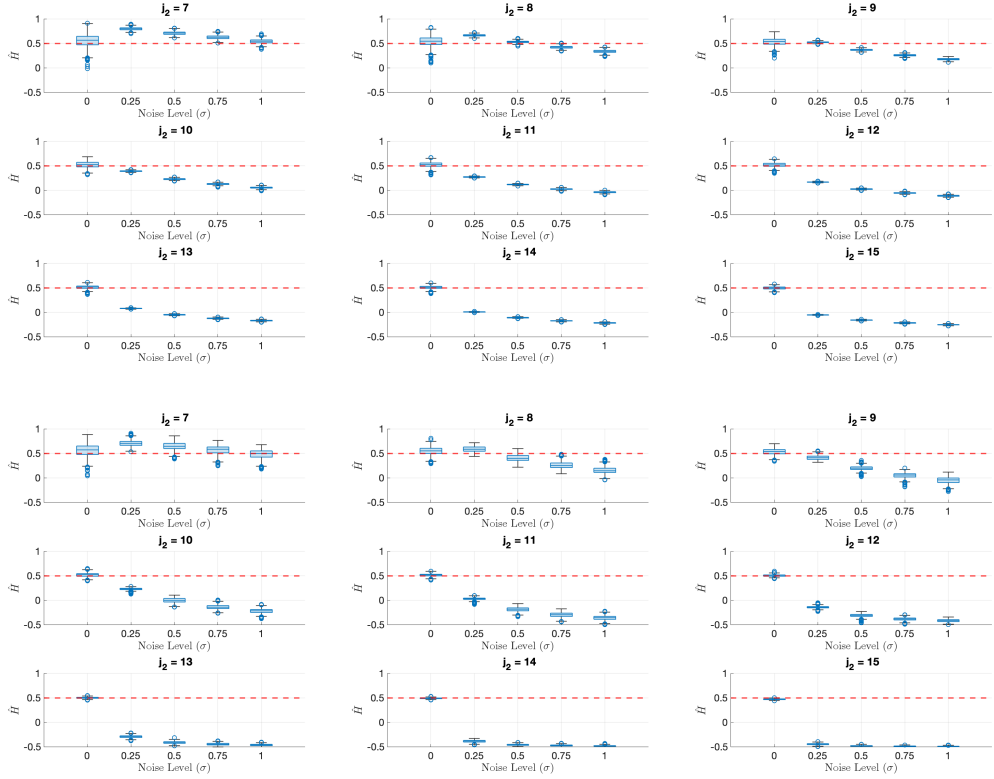

Figure 10: Analysis of the noise level ($\sigma_\epsilon$) and scale range on the estimation of the Hurst exponent ($\hat{H}$) using the Standard and ALPHEE methods. The true Hurst exponent ($H = 0.5$) is shown as a red dashed line, with the lower scale fixed at $j_1 = 3$ and the upper scale $j_2$ varying from 7 to 15.